\documentclass[10pt,twocolumn,letterpaper]{article}

\usepackage{cvpr}
\usepackage{times}
\usepackage{epsfig}
\usepackage{graphicx}
\usepackage{amsmath}
\usepackage{amssymb}
\usepackage{algorithm}
\usepackage{algpseudocode}
\usepackage{booktabs}       
\usepackage{mathtools}
\usepackage{multirow}
\usepackage{placeins}
\usepackage{siunitx}
\usepackage{subcaption}

\DeclarePairedDelimiter\floor{\lfloor}{\rfloor}%
\usepackage[pagebackref=true,breaklinks=true,letterpaper=true,colorlinks,bookmarks=false]{hyperref}

\cvprfinalcopy 

\def\cvprPaperID{****} 

\ifcvprfinal\fi
\begin{document}

\title{\vspace{-2em}Nail Polish Try-On: Realtime Semantic Segmentation of Small Objects for
       Native and Browser Smartphone AR Applications\vspace{-.5em}}

\author{%
Brendan Duke \hspace{9pt} Abdalla Ahmed \hspace{9pt} Edmund Phung \hspace{9pt}
Irina Kezele \hspace{9pt} Parham Aarabi\\
\small{ModiFace Inc.}\\
{\tt\small \{brendan,abdalla,edmund,irina,parham\}@modiface.com}
}

\maketitle

\begin{abstract}
\vspace{-.5em}
We provide a system for semantic segmentation of small objects that enables
nail polish try-on AR applications to run client-side in realtime in native and
web mobile applications.
By adjusting input resolution and neural network depth, our model design enables
a smooth trade-off of performance and runtime, with the highest performance
setting achieving~\num{94.5} mIoU at 29.8ms runtime in native applications on
an iPad Pro.
We also provide a postprocessing and rendering algorithm for nail polish
try-on, which integrates with our semantic segmentation and fingernail base-tip
direction predictions.
Project page:~\url{https://research.modiface.com/nail-polish-try-on-cvprw2019}
\end{abstract}
\vspace{-1em}

\vspace{-1em}
\section{Introduction}

We present our end-to-end solution for simultaneous realtime tracking of
fingernails and rendering of nail polish.
Our system locates and identifies fingernails from a videostream in realtime
with pixel accuracy, and provides enough landmark information, e.g.,
directionality, to support rendering techniques.

We collected an entirely new dataset with semantic segmentation and landmark
labels, developed a high-resolution neural network model for mobile devices,
and combined these data and this model with postprocessing and rendering
algorithms for nail polish try-on.

We deployed our trained models on two hardware platforms: iOS via CoreML, and
web browsers via TensorFlow.js~\cite{smilkov2019tensorflowjs}.
Our model and algorithm design is flexible enough to support both the higher
computation native iOS platform, as well as the more resource constrained web
platform, by making only minor tweaks to our model architecture, and without
any major negative impact on performance.

\subsection{Contributions}

Below we enumerate our work's novel contributions.
\vspace{-.5em}

\begin{itemize}
        \item We created a dataset of~\num{1438} images sourced from photos
                and/or videos of the hands of over~\num{300} subjects, and
                annotated with foreground-background, per-finger class, and
                base-tip direction field labels.
\vspace{-.5em}

        \item We developed a novel neural network architecture for semantic
                segmentation designed for both running on mobile devices and
                precisely segmenting small objects.
\vspace{-.5em}

        \item We developed a postprocessing algorithm that uses multiple
                outputs from our fingernail tracking model to both segment
                fingernails and localize individual fingernails, as well as to
                find their 2D orientation.
\vspace{-.5em}
\end{itemize}

\section{Related Work}

Our work builds on MobileNetV2~\cite{sandler2018mobilenetv2} by using it as an
encoder-backbone in our cascaded semantic segmentation model architecture.
However, our system is agnostic to the specific encoder model used, so any
existing efficient model from the
literature~\cite{forrest2016squeezenet, wang2018pelee, zhang2018shufflenet, zoph2018learning}
could be used as a drop-in replacement for our encoder.

Our Loss Max-Pooling (LMP) loss is based on~\cite{bulo2017loss}, where we fix
their~$p$-norm parameter to~$p = 1$ for simplicity.
Our experiments further support the effectiveness of LMP in the intrinsically
class imbalanced fingernail segmentation task.

Our cascaded architecture is related to ICNet~\cite{zhao2017icnet} in
that our neural network model combines shallow/high-resolution and
deep/low-resolution branches.
Unlike ICNet, our model is designed to run on mobile devices, and therefore we
completely redesigned the encoder and decoder based on this requirement.

\section{Methods}

\subsection{Dataset}

Due to a lack of prior work specifically on fingernail tracking, we created an
entirely new dataset for this task.
We collected egocentric data from participants, who we asked to take either
photos or videos of their hands as if they were showing off their fingernails
for a post on social media.

Our annotations included a combination of three label types:
foreground/background polygons, individual fingernail class polygons, and
base/tip landmarks to define per-polygon orientation.
We used the fingernail base/tip landmarks to generate a dense direction field,
where each pixel contains a base-to-tip unit vector for the fingernail that
that pixel belongs to.

Our annotated dataset consists of~\num{1438} annotated images in total, which
we split into train, validation and test sets based on the participant who
contributed the images.
The split dataset contains~\num{941}, \num{254} and \num{243} images each in
train, val and test, respectively.

\subsection{Model}

\begin{figure}[t]
\centering
\includegraphics[width=0.9\linewidth]{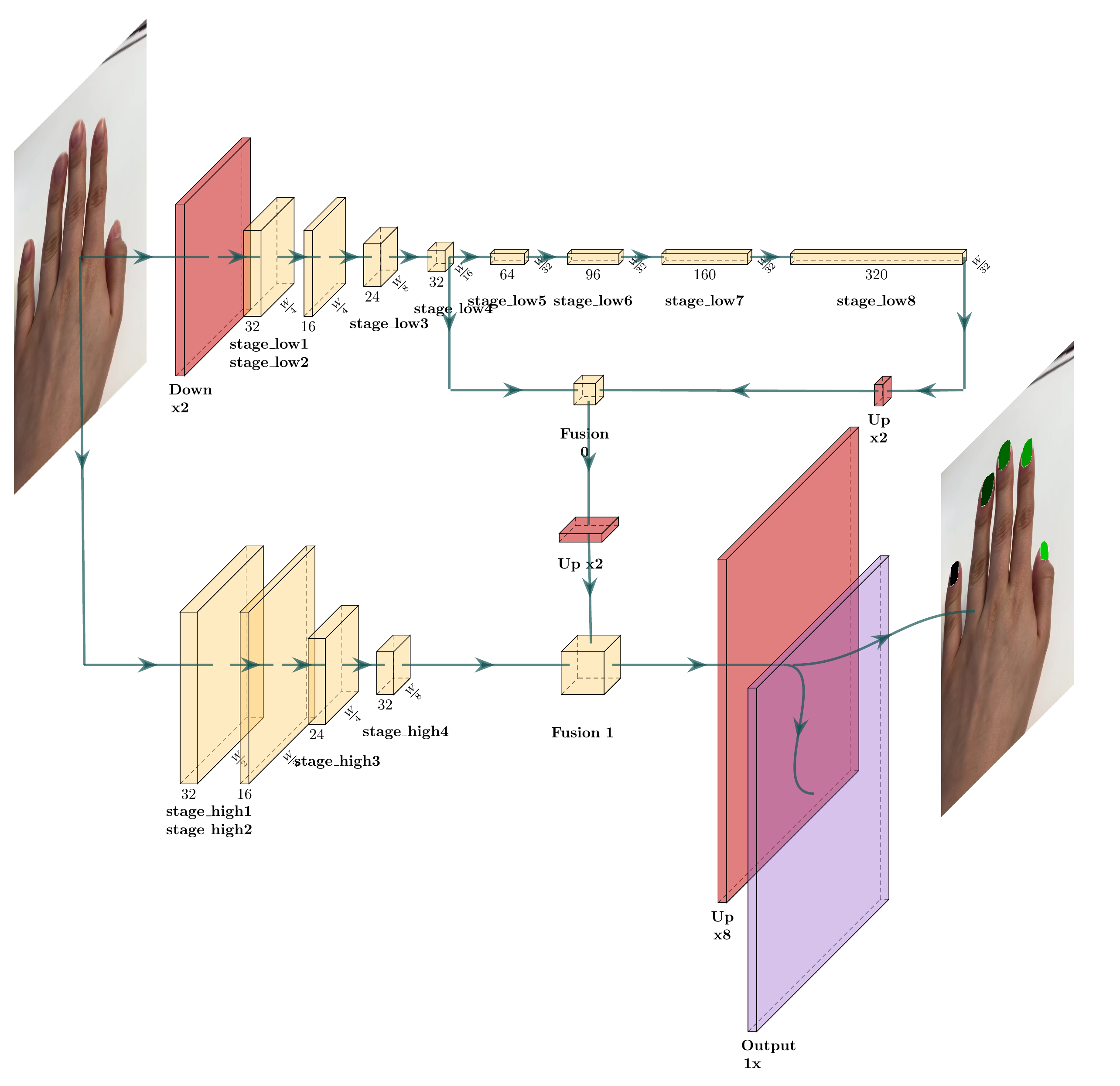}
\vspace{-1em}
\caption{A top level overview of our nail segmentation model.
         In this example, the input resolution is~$H\times W$.
         See Figure~\ref{fig:modelcomponents} for fusion and output block
         architectures.}
\vspace{-1em}
\label{fig:model-top}
\end{figure}

\begin{figure}[t]
\begin{subfigure}{0.5\linewidth}
\centering
\includegraphics[width=1.0\linewidth]{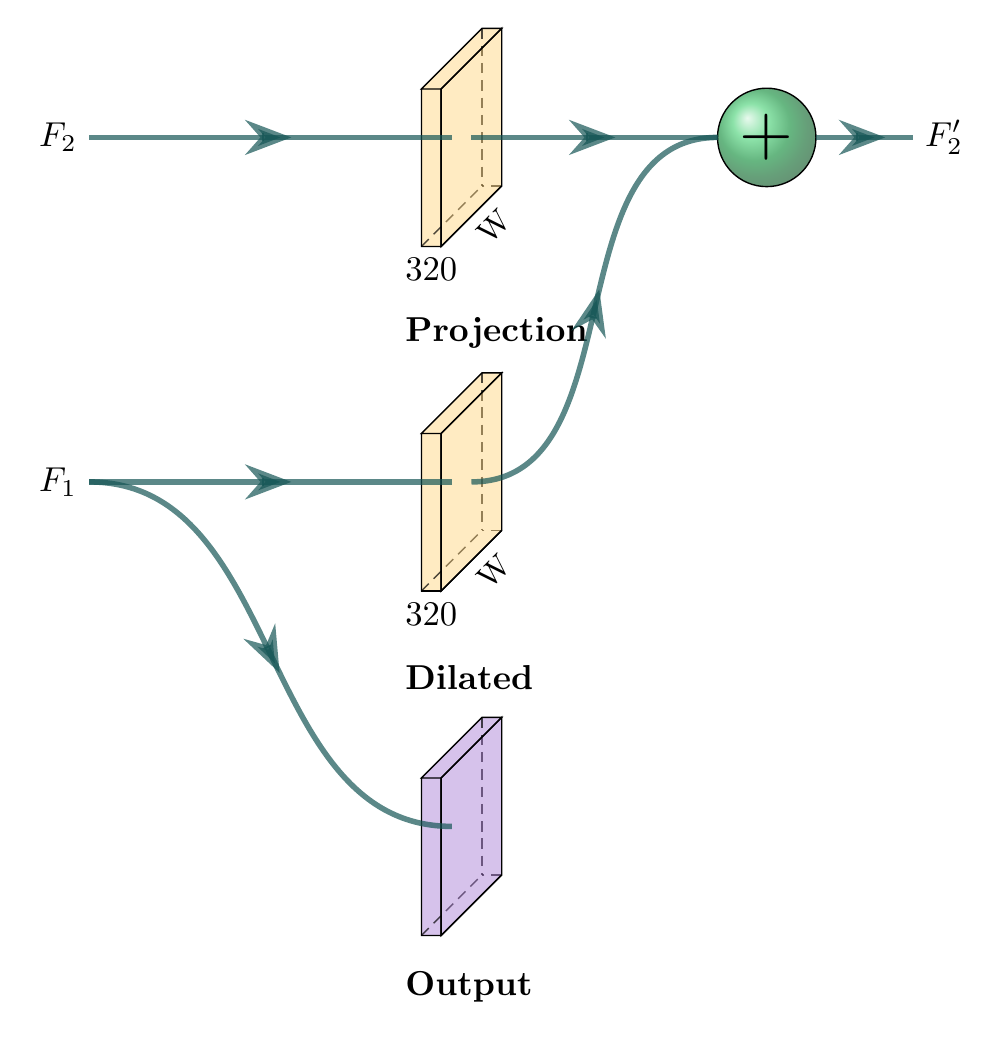}
\end{subfigure}%
\begin{subfigure}{0.5\linewidth}
\centering
\includegraphics[width=1.0\linewidth]{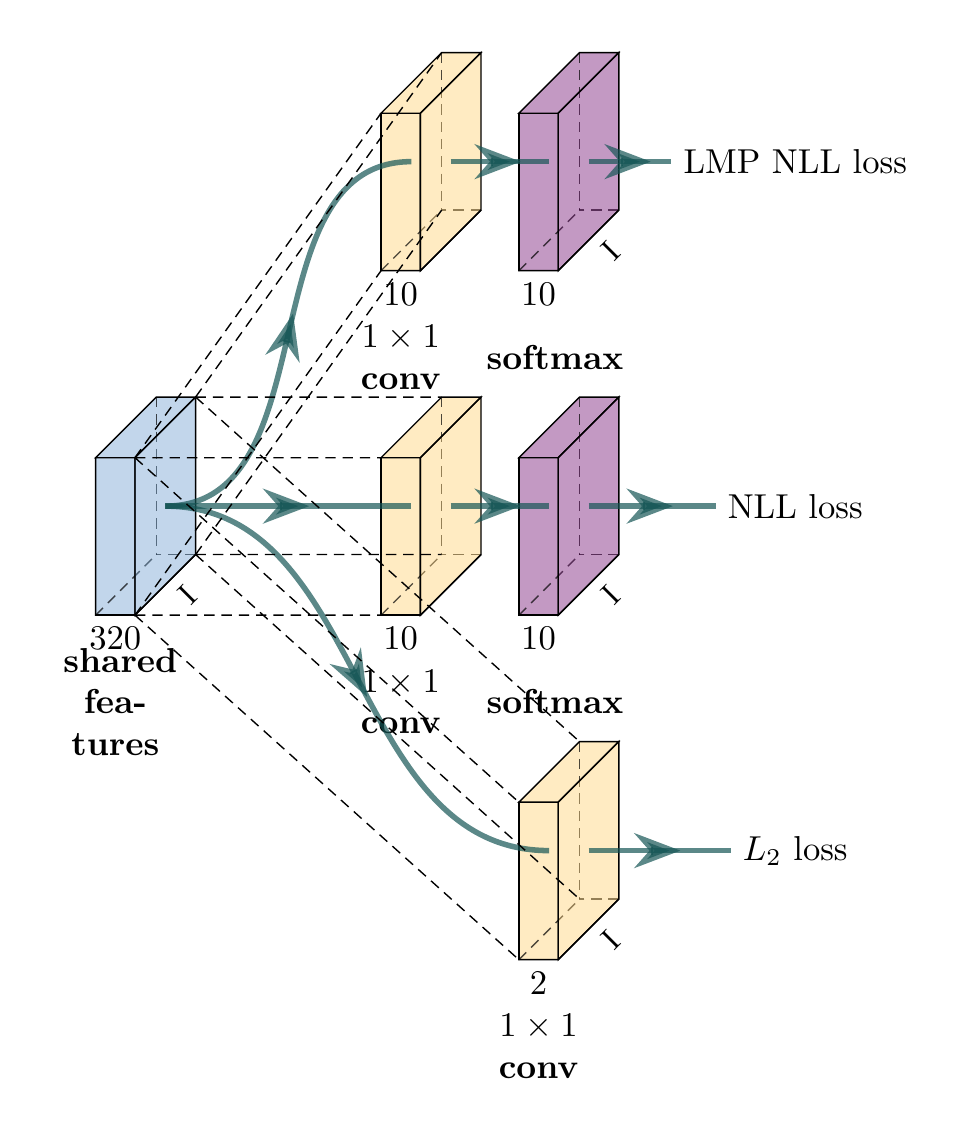}
\end{subfigure}
\vspace{-1.5em}
\caption{The fusion module (left) and output branch (right) of our model.
         We fuse upsampled low-resolution, deep features~$F_1$ with
         high-resolution, shallow features~$F_2$ to produce high-resolution
         fused features~$F_2'$ in the decoder.
         The output branch is used in each fusion block, and at the
         full-resolution output layer of the model.}
\vspace{-2em}
\label{fig:modelcomponents}
\end{figure}

The core of our nail tracking system is an encoder-decoder convolutional neural
network~(CNN) architecture trained to output foreground/background and
fingernail class segmentations, as well as base-tip direction fields.
Our model architecture draws inspiration from ICNet~\cite{zhao2017icnet}, which
we improved to meet the runtime and size constraints of mobile devices, and to
produce our multi-task outputs.
A top-level view of our model architecture is in Figure~\ref{fig:model-top}.

We initialized the encoder of our model with
MobileNetV2~\cite{sandler2018mobilenetv2} model weights pretrained on
ImageNet~\cite{deng2009imagenet}.
We used a cascade of two~$\alpha = 1.0$ MobileNetV2 encoder backbones, both
pretrained on~$224\times224$ ImageNet images.
The encoder cascade consists of one shallow network with high-resolution inputs
(\verb|stage_high1..4|), and one deep network with low-resolution inputs
(\verb|stage_low1..8|), both of which are prefixes of the full MobileNetV2.
To maintain a higher spatial resolution of feature maps output by the
low-resolution encoder we changed stage 6 from stride 2 to stride 1, and used
dilated~$2\times$ convolutions in stages~7 and~8 to compensate, reducing its
overall stride from~$32\times$ to~$16\times$.

Our model's decoder is shown in the middle and bottom right of
Figure~\ref{fig:model-top}, and a detailed view of the decoder fusion block is
shown in Figure~\ref{fig:modelcomponents}.
For an original input of size~$H\times W$, our decoder fuses
the~$\frac{H}{16}\times\frac{W}{16}$ features from~\verb|stage_low4| with the
upsampled features from~\verb|stage_low8|, then upsamples and fuses the
resulting features with the~$\frac{H}{8}\times\frac{W}{8}$ features
from~\verb|stage_high4|.
The convolution on~$F_1$ is dilated to match~$F_1$'s~$\times 2$ upsampling
rate, and the elementwise sum (to form~$F_2'$) allows the projected~$F_2$ to
directly refine the coarser~$F_1$ feature maps.

As shown in Figure~\ref{fig:modelcomponents}, a~$1\times1$ convolutional classifier is
applied to the upsampled~$F_1$ features, which are used to predict downsampled
labels.
As in~\cite{ghiasi2016laplacian}, this ``Laplacian pyramid'' of outputs
optimizes the higher-resolution, smaller receptive field feature maps to focus
on refining the predictions from low-resolution, larger receptive field feature
maps.

Also shown in Figure~\ref{fig:modelcomponents}, our system uses multiple output
decoder branches to provide directionality information (i.e., vectors from base
to tip) needed to render over fingernail tips, and fingernail class predictions
needed to find fingernail instances using connected components.
We trained these additional decoders to produce dense predictions penalized
only in the annotated fingernail area of the image.

\subsection{Criticism}

To deal with the class imbalance between background (overrepresented class) and
fingernail (underrepresented class), in our objective function we used
LMP~\cite{bulo2017loss} over all pixels in a minibatch by sorting by the loss
magnitude of each pixel, and taking the mean over the top~\num{10}\% of pixels
as the minibatch loss.
Compared with a baseline that weighted the fingernail class's loss
by~$20\times$ more than the background, LMP yielded a gain
of~$\approx 2\%$~mIoU, reflected in sharper nail edges where the baseline
consistently oversegmented.

We used three loss functions corresponding to the three outputs of our model
shown in Figure~\ref{fig:modelcomponents}.
The fingernail class and foreground/background predictions both minimize the
negative log-likelihood of a multinomial distribution given in
Equation~\ref{eqn:nll}, where~$c$ is the ground truth class and~$x_c^{ij}$ is the
pre-softmax prediction of the~$c$th class by the model.
Each of the following equations refers to indices~$i$ and~$j$ corresponding to
pixels at~$(x, y) = (i, j)$.

\vspace{-.75em}
\begin{equation}
\mathcal{L}_{class,fgbg}^{ij} = -\log\left(\frac{\exp x_c^{ij}}{\sum_{c'}\exp x_{c'}^{ij}}\right)
\label{eqn:nll}
\end{equation}
\vspace{-.75em}

In the case of class predictions,~$c \in \{1, 2, \dots, 10\}$, while for
foreground/background predictions~$c \in \{1, 2\}$.
We used LMP for foreground/background predictions only; since fingernail class
predictions are only valid in the fingernail region, those classes are
balanced and do not require LMP\@.

\vspace{-.75em}
\begin{equation}
\ell_{fgbg} = \frac{1}{N_{thresh}} \sum_{ij} \left[\mathcal{L}_{fgbg}^{ij} > \tau\right]\cdot \mathcal{L}_{fgbg}^{ij}
\label{eqn:fgbgloss}
\end{equation}
\vspace{-.75em}

In
Equation~\ref{eqn:fgbgloss},~$N_{thresh} = \sum_{ij} \left[\mathcal{L}_{fgbg}^{ij} > \tau\right]$,
and threshold~$\tau$ is the loss value of the~$\floor{0.1\times H\times W}$th
highest loss pixel. The~$[\cdot]$ operator is the indicator function.

For the direction field output, we apply an~$L_2$ loss (other loss functions
such as~Huber and~$L_1$ could be used in its place) on the normalized base to
tip direction of the nail for each pixel inside the ground truth nail.

\vspace{-.75em}
\begin{equation}
 \ell_{field} = \frac{1}{H\times W}\sum_{ij} {\lVert\hat{\mathbf{u}}_{field}^{ij} - \mathbf{u}_{field}^{ij}\rVert}_2^2
\label{eqn:fieldloss}
\end{equation}
\vspace{-.75em}

The field direction labels are normalized so
that~${\lVert \mathbf{u}_{field}^{ij} \rVert}_2 = 1$.
Since the direction field and the fingernail class losses have no class
imbalance problem, we simply set these two losses to the means of their
respective individual losses (i.e.,
$\ell_{class} = \frac{1}{H\times W}\sum_{ij} \mathcal{L}_{class}^{ij}$).
The overall loss is~$\ell = \ell_{fgbg} + \ell_{class} + \ell_{field}$.

\subsection{Postprocessing and Rendering}

Our postprocessing and rendering algorithm uses the output of our tracking
predictions to draw realistic nail polish on a user's fingernails.
Our algorithm uses the individual fingernail location and direction information
predicted by our fingernail tracking module in order to render gradients, and
to hide the light-coloured distal edge of natural nails.

We first use connected components~\cite{grana2010optimized} on the class
predictions to extract a connected blob for each nail.
We then estimate each nail's angle by using the foreground (nail) softmax
scores to take a weighted average of the direction predictions in each nail
region.
We use the fingernails' connected components and orientations to implement
rendering effects, including rendering gradients to approximate specular
reflections, and stretching the nail masks towards the fingernail tips in order
to hide the fingernails' light-coloured distal edges.

\subsection{Performance and Runtime Results}

We trained our neural network models using PyTorch~\cite{paszke2017automatic}.
We deployed the trained models to iOS using CoreML, and to web browsers using
TensorFlow.js~\cite{smilkov2019tensorflowjs}.

Given current software implementations, namely CoreML and TensorFlow.js, and
current mobile device hardware, our system could run in realtime (i.e.,
at~$\geq 10$ FPS) at all resolutions up to~$640\times480$ (native mobile)
and~$480\times360$ (web mobile), for which we trained on randomly-cropped input
resolutions of~$448\times448$ and~$336\times336$, respectively.

\begin{figure}[t]
\begin{subfigure}{0.3\linewidth}
 \small
\centering
\begin{tabular}{l|c}
 {\bf Model} &  mIoU  \\
 \midrule
 Baseline &  91.3 \\
 + LMP &  93.3 \\
 + Cascade &  94.5 \\
\end{tabular}
\end{subfigure}%
\hspace{.5em}
\begin{subfigure}{0.7\linewidth}
\includegraphics[width=\linewidth]{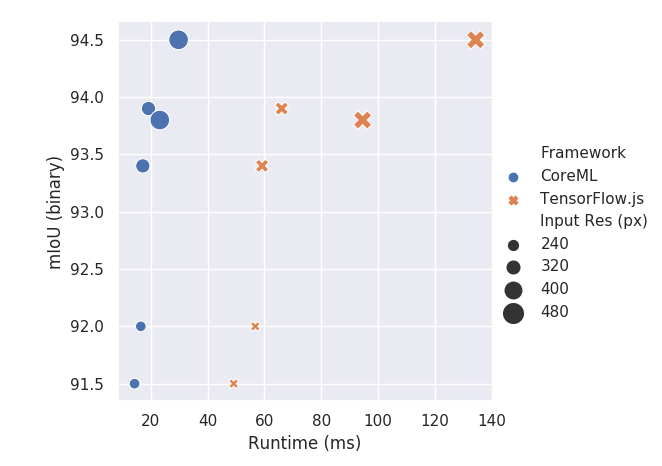}
\end{subfigure}%
\vspace{-1em}
\caption{Runtime/mIoU comparison for model settings on an iPad Pro device
 (right) and ablation study of LMP and cascade model architecture features
 (left).}
\vspace{-1em}
\label{fig:perfruntime}
\end{figure}

In Figure~\ref{fig:perfruntime} we evaluate performance in binary mIoU on our
validation set against runtime on an iPad Pro device on CoreML (native iOS) and
TensorFlow.js (Safari web browser) software platforms using two variants of our
model (55-layer and 43-layer encoders) at three resolutions (288px, 352px, and
480px), and compare the effect of introducing LMP and the low/high-resolution
cascade into our model design.

\section{Workshop Demo}

For our submission we propose both a web browser-based and iOS demo of our
virtual nail polish try-on technology.
We will provide an iOS tablet device for users to try the virtual nail try-on
application in person, as well as a web link
at~\url{https://ola.modiface.com/nailsweb/cvpr2019demo} (best viewed on a
mobile device using Safari on iOS, Chrome or Firefox on Android) that users can
access with their own devices and use the technology by enabling camera access
in the browser (``https'' prefix required).

\FloatBarrier{}

\vspace{-0.5em}
\section{Conclusion}

We presented a fingernail tracking system that runs in realtime on both iOS and
web platforms, and enables pixel-accurate fingernail predictions at up
to~$640\times480$ resolution.
We also proposed a postprocessing and rendering algorithm that integrates with
our model to produce realistic nail polish rendering effects.

{\small
\bibliographystyle{ieee_fullname}
\bibliography{nails-extended-abstract}

\begin{thebibliography}{10}\itemsep=-1pt

\bibitem{bulo2017loss}
Samuel~Rota Bul{\`{o}}, Gerhard Neuhold, and Peter Kontschieder.
\newblock Loss max-pooling for semantic image segmentation.
\newblock In {\em The IEEE Conference on Computer Vision and Pattern
  Recognition (CVPR)}, 2017.

\bibitem{deng2009imagenet}
J. Deng, W. Dong, R. Socher, L.-J. Li, K. Li, and L. Fei-Fei.
\newblock {ImageNet: A Large-Scale Hierarchical Image Database}.
\newblock In {\em The IEEE Conference on Computer Vision and Pattern
  Recognition (CVPR)}, 2009.

\bibitem{ghiasi2016laplacian}
Golnaz Ghiasi and Charless~C. Fowlkes.
\newblock Laplacian reconstruction and refinement for semantic segmentation.
\newblock In {\em ECCV}, 2016.

\bibitem{grana2010optimized}
C. Grana, D. Borghesani, and R. Cucchiara.
\newblock Optimized block-based connected components labeling with decision
  trees.
\newblock {\em IEEE Transactions on Image Processing}, 2010.

\bibitem{forrest2016squeezenet}
Forrest~N. Iandola, Song Han, Matthew~W. Moskewicz, Khalid Ashraf, William~J.
  Dally, and Kurt Keutzer.
\newblock Squeezenet: Alexnet-level accuracy with 50x fewer parameters and
  $<$0.5mb model size.
\newblock {\em arXiv:1602.07360}, 2016.

\bibitem{paszke2017automatic}
Adam Paszke, Sam Gross, Soumith Chintala, Gregory Chanan, Edward Yang, Zachary
  DeVito, Zeming Lin, Alban Desmaison, Luca Antiga, and Adam Lerer.
\newblock Automatic differentiation in pytorch.
\newblock In {\em NIPS-W}, 2017.

\bibitem{sandler2018mobilenetv2}
Mark Sandler, Andrew Howard, Menglong Zhu, Andrey Zhmoginov, and Liang-Chieh
  Chen.
\newblock Mobilenetv2: Inverted residuals and linear bottlenecks.
\newblock In {\em The IEEE Conference on Computer Vision and Pattern
  Recognition (CVPR)}, June 2018.

\bibitem{smilkov2019tensorflowjs}
Daniel Smilkov, Nikhil Thorat, Yannick Assogba, Ann Yuan, Nick Kreeger, Ping
  Yu, Kangyi Zhang, Shanqing Cai, Eric Nielsen, David Soergel, Stan Bileschi,
  Michael Terry, Charles Nicholson, Sandeep~N. Gupta, Sarah Sirajuddin, D.
  Sculley, Rajat Monga, Greg Corrado, Fernanda~B. Vi{\'{e}}gas, and Martin
  Wattenberg.
\newblock Tensorflow.js: Machine learning for the web and beyond.
\newblock {\em arXiv preprint arXiv:1901.05350}, 2019.

\bibitem{wang2018pelee}
Robert~J Wang, Xiang Li, and Charles~X Ling.
\newblock Pelee: A real-time object detection system on mobile devices.
\newblock In {\em Advances in Neural Information Processing Systems 31}, 2018.

\bibitem{zhang2018shufflenet}
Xiangyu Zhang, Xinyu Zhou, Mengxiao Lin, and Jian Sun.
\newblock Shufflenet: An extremely efficient convolutional neural network for
  mobile devices.
\newblock In {\em The IEEE Conference on Computer Vision and Pattern
  Recognition (CVPR)}, 2018.

\bibitem{zhao2017icnet}
Hengshuang Zhao, Xiaojuan Qi, Xiaoyong Shen, Jianping Shi, and Jiaya Jia.
\newblock Icnet for real-time semantic segmentation on high-resolution images.
\newblock In {\em ECCV}, 2018.

\bibitem{zoph2018learning}
Barret Zoph, Vijay Vasudevan, Jonathon Shlens, and Quoc~V. Le.
\newblock Learning transferable architectures for scalable image recognition.
\newblock In {\em The IEEE Conference on Computer Vision and Pattern
  Recognition (CVPR)}, 2018.

\end{thebibliography}
}

\end{document}